%% file: siggraph.tex
\documentclass[acmtog]{acmart}
\copyrightyear{2026}
\acmYear{2026}
\setcopyright{cc}
\setcctype{by}
\acmConference[SIGGRAPH Conference Papers '26]{Special Interest Group on Computer Graphics and Interactive Techniques Conference Conference Papers}{July 19--23, 2026}{Los Angeles, CA, USA}
\acmBooktitle{Special Interest Group on Computer Graphics and Interactive Techniques Conference Conference Papers (SIGGRAPH Conference Papers '26), July 19--23, 2026, Los Angeles, CA, USA}
\acmDOI{10.1145/3799902.3811174}
\acmISBN{979-8-4007-2554-8/2026/07}

\input{src/macros}

\makeatletter
\renewcommand{\@fnsymbol}[1]{\ensuremath{\ifcase#1\or \dagger\or *\or \ddagger\or
   \mathsection\or \mathparagraph\or \|\or **\or \dagger\dagger
   \or \ddagger\ddagger \else\@ctrerr\fi}}
\makeatother

\begin{document}
\title{\changed{Img2CADSeq: Image-to-CAD Generation via Sequence-Based Diffusion}}

\author{Shiyu Tan}
\email{tansy25@mails.tsinghua.edu.cn}
\affiliation{%
	\department{School of Software}
	\institution{Tsinghua University}
	\country{China}	
}

\author{Zixuan Zhao}
\authornote{Authors with equal contribution.} 
\email{1449465595@qq.com}
\affiliation{%
	\department{School of Software}
	\institution{Tsinghua University}
	\country{China}	
}

\author{Hao Gao}
\authornotemark[1] 
\email{hgao1743@gmail.com}
\affiliation{%
	\department{School of Software}
	\institution{Tsinghua University}
	\country{China}	
}

\author{Zhiheng Chen}
\authornotemark[1] 
\email{chenzhih24@mails.tsinghua.edu.cn}
\affiliation{%
	\department{School of Software}
	\institution{Tsinghua University}
	\country{China}	
}

\author{Xiaolong Yin}
\email{1505631565@qq.com}
\affiliation{%
	\department{School of Software}
	\institution{Tsinghua University}
	\country{China}	
}

\author{Enya Shen}
\authornote{Corresponding author.} 
\email{shenenya@tsinghua.edu.cn}
\affiliation{%
	\department{School of Software}
	\institution{Tsinghua University}
	\country{China}	
}

\begin{abstract}

Boundary Representation (BRep) is the standard format for Computer-Aided Design (CAD), yet reconstructing high-quality BReps from single-view images remains challenging due to the complexity of topological constraints and operation sequences. We present Img2CADSeq, a \changed{multi-stage} pipeline that overcomes these limitations by encoding CAD sequences into a three-level hierarchical codebook. \changed{Guided by an importance prioritization}, this strategy values profiles over details, compressing long sequences into a stable discrete latent space. To bridge the modality gap, we leverage a coarse-to-fine point cloud intermediate, aligning 2D visual features with 3D CAD sequences via contrastive learning to condition a VQ-Diffusion model. Supported by newly introduced CAD-220K and PrintCAD datasets, our approach ensures robust industrial domain adaptation. Extensive experiments demonstrate that Img2CADSeq significantly outperforms state-of-the-art methods, producing standard STEP files that can be directly used in commercial CAD software.
\changed{Code and data for this paper are at \url{https://github.com/Rilpraa0110/Img2CADSeq}}
\end{abstract}



\begin{teaserfigure}
  \centering
  \includegraphics[width=\linewidth]{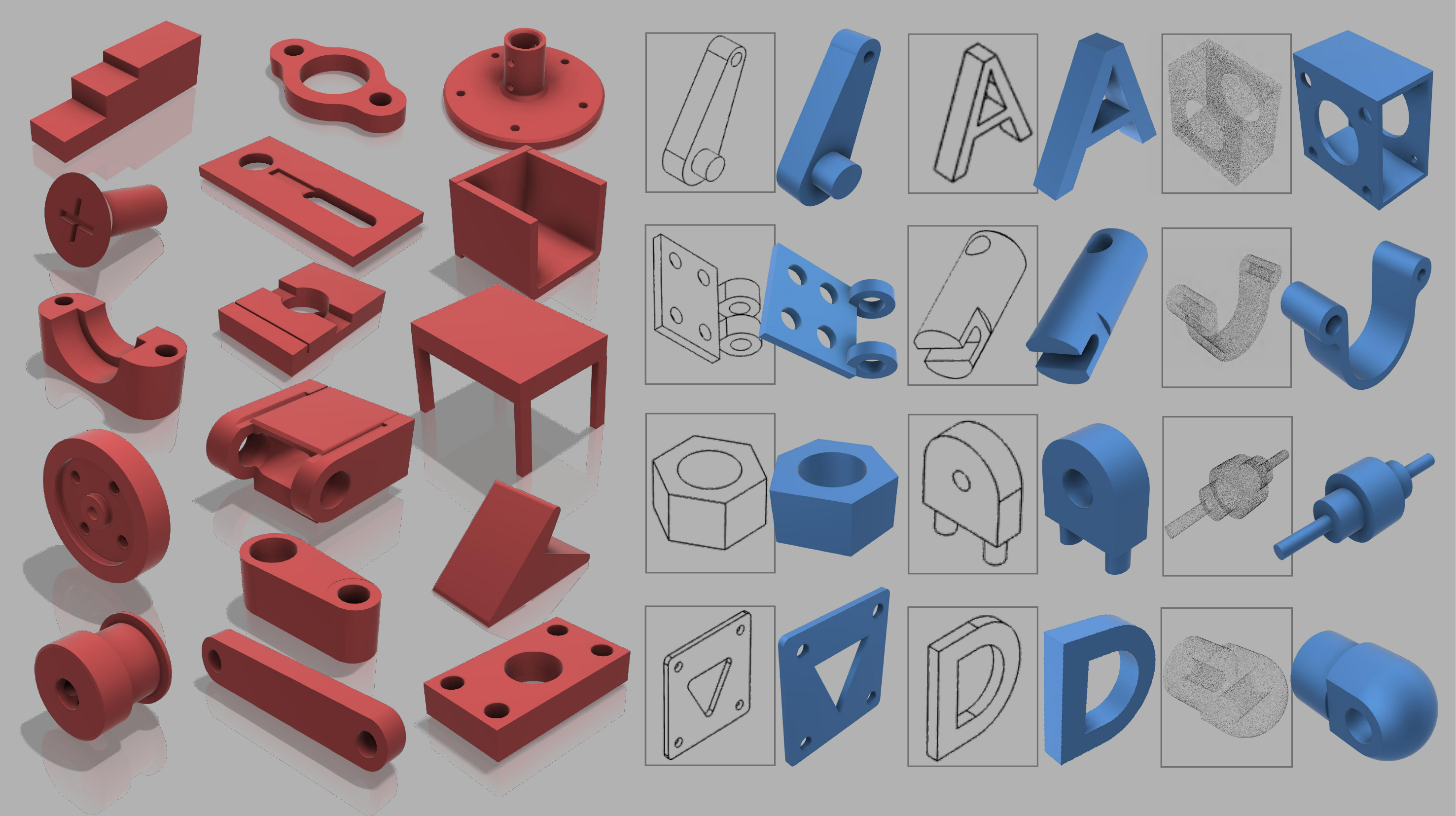}
  \caption{
Our proposed {\em \changed{Img2CADSeq} }  is a novel method based on boundary representations (BReps) structure. It is \changed{a multi-stage} pipeline that can generate standardized STEP files. The fourth and fifth columns show reconstructed results generated with single-view image conditioning. The method also delivers strong results in unconditional generation, as seen in the first three columns with red parts and cloud-conditioned generation is shown in the sixth column. It surpasses existing state-of-the-art models in mechanical components generation.
  }
  \label{fig:teaser}
\end{teaserfigure}

\maketitle

\section{Introduction}
\label{sec:intro}

In recent years, single-view 3D generation has witnessed remarkable progress, driven by advancements in differentiable rendering techniques such as Neural Radiance Fields (NeRF)~\cite{mildenhall2021nerf} and Gaussian Splatting~\cite{kerbl20233d}. While these methods excel at synthesizing visually detailed 3D shapes from single-view images, the resulting representations are typically implicit fields or unstructured point splats. Crucially, they lack the clear internal topology of geometric curves and surfaces required for downstream applications~\cite{lin2024dynamic,kim20243dgs,rossignac2002csg}. Consequently, they are ill-suited for precision-demanding tasks such as industrial design and manufacturing, raising a fundamental question: How can we generate 3D data that is not only visually coherent but also editable and compatible with engineering standards?

Boundary Representation (BRep) serves as the standard format for Computer-Aided Design (CAD)~\cite{miyazaki2009review}, explicitly describing objects via precise parametric geometry and topology. However, reconstructing high-quality BRep models from single images remains a significant challenge~\cite{kasik2005ten}. Unlike tensor-based formats favored by neural networks, BRep relies on complex topological relationships between geometric entities. While some studies~\cite{wu2021deepcad} attempt to encode BReps as sequences of construction operations, these sequences are often excessively long and unstructured, making them difficult for deep models to learn effectively. To address these challenges and bridge the gap between pixel-level vision and parameter-level engineering, we present Img2CADSeq (see Fig. \ref{fig:teaser}), a multi-stage framework composed of three key innovations.

First, we tackle the representation difficulty with a hierarchical encoding strategy. Inspired by importance prioritization design processes~\cite{visser2006designing}, where designers favor ``overall profiles'' before detailing ``local features'', we propose a three-layer codebook. This approach quantizes CAD operation sequences into three levels: Curve-Cluster, Sketch-Patch, and Extrude-Block.
This approach naturally captures the design intent and specific parameters. Furthermore, the sequences have a sequential structure similar to natural language, which makes them easier for neural models to learn. 
Meanwhile, by focusing on relative geometric features rather than absolute coordinates, this structure significantly compacts the sequence length while implicitly preserving topological validity.

 Second, to bridge the large semantic gap between 2D images and BRep sequences, we introduce a coarse-to-fine point cloud intermediate representation. We start by employing Dens3R~\cite{fang2025dens3r} to lift the single-view image into an initial coarse point cloud. This approach was chosen for the flexibility of its QV linear layers, which allow for efficient training and optimization. However, a critical bottleneck is that existing single-view to point-cloud models are typically trained on non-industrial datasets (e.g., ShapeNet~\cite{chang2015shapenet}), which fail to capture the geometric rigor of mechanical parts. To solve this, we \changed{provide two large-scale datasets: CAD-220K, a curated subset of ABC dataset comprising over 220,000 diverse 3D CAD models paired with rendered images and other modality forms. PrintCAD, a dataset of over 2,000 3D-printed components images captured under real-world lighting conditions and backgrounds}. By training on these datasets, we obtain a robust coarse geometry. Subsequently, to mitigate the inherent noise in the generated geometry and capture high-frequency details, we process this output using a novel Uncertainty-Aware DGCNN (UA-DGCNN). This module leverages importance estimation to weigh reliable features, utilizing \changed{heuristic-guided resampling} to extract a robust fine-grained encoding for the subsequent alignment task.

\begin{figure*}
  \centering
  \includegraphics[width=\linewidth]{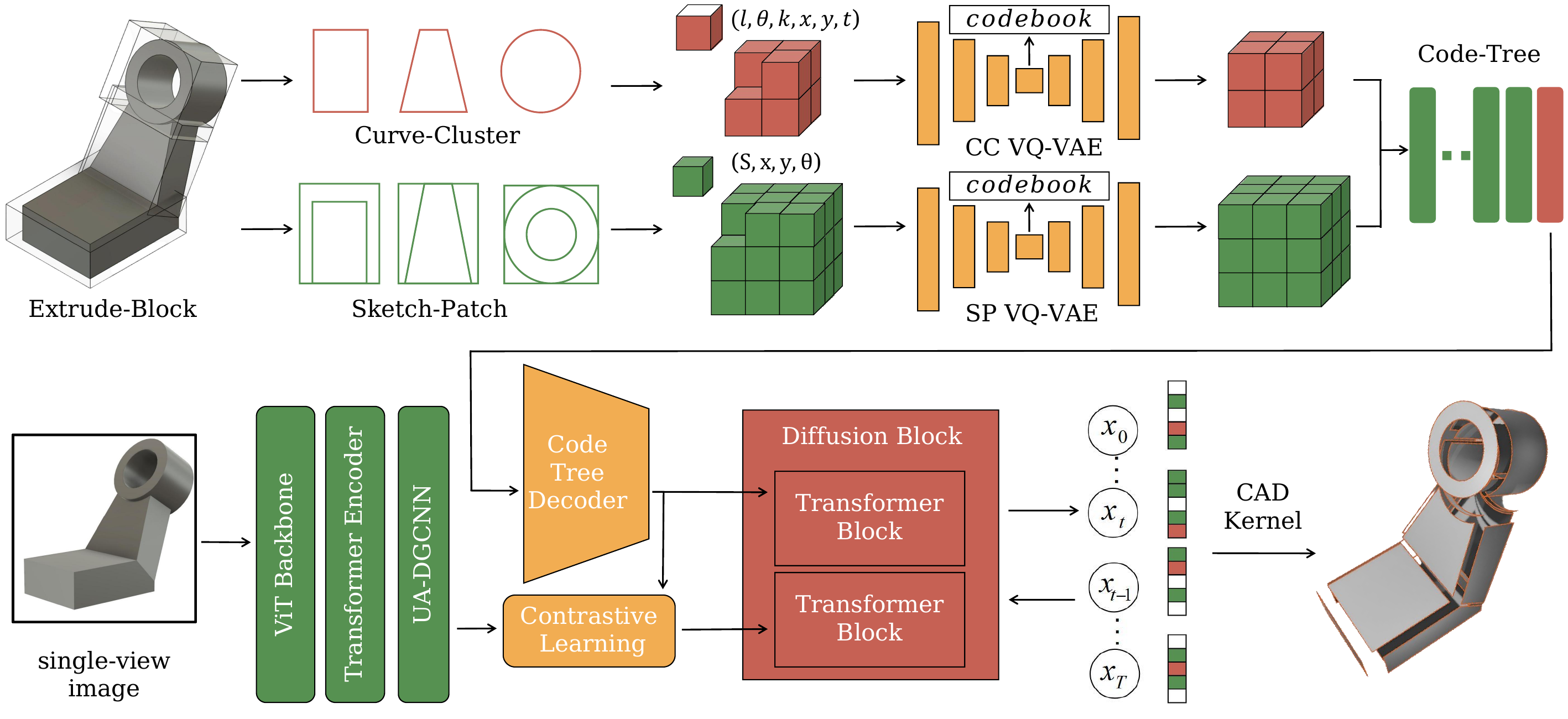}
  \caption{
\textbf{Overview of the Img2CADSeq Framework.} In the first stage, hierarchical sequence encoding represents CAD operations via a three-level codebook into a discrete space.
 Then we lift the input image into a 3D point cloud using a tailored network trained jointly on both synthetic and real-world data types, which is then refined by UA-DGCNN to sharpen edges and smooth surfaces. Finally, we employ contrastive learning to align the geometric embeddings with the CAD latent space, guiding a VQ-Diffusion model to predict a valid CAD operation sequence to be compiled into a watertight BRep.
  }
  \label{fig:pipeline}
\end{figure*}

Third, we propose a robust alignment mechanism to guide generation. Merely having a point cloud is insufficient; we must semantically connect it to our CAD codebook. We employ contrastive learning to align the features of the generated point cloud with the latent space of the CAD operation sequences. This alignment ensures that the structural information from the point cloud is effectively translated into the CAD language. Finally, these features are injected as a condition into a VQ-Diffusion model~\cite{gu2022vector}, which generates the discrete tokens for the CAD sequence. A geometric kernel then compiles these tokens into the final BRep model (see Fig. \ref{fig:pipeline}).

In summary, the key contributions of this work are as follows:
\begin{itemize}

\item We propose a hierarchical three-layer codebook to encode CAD operation sequences. Following a "profile-to-detail" logic, this strategy compacts complex sequences into a discrete latent space suitable for diffusion modeling.

\item \changed{To address the scarcity of industrial data, a key contribution of our work lies in the  combination of two distinct data types: curated synthetic models (CAD-220K) and real-world captured objects (PrintCAD). By leveraging these two joint data types, we train a network to generate intermediate point clouds. This combination explicitly bridges the sim-to-real gap, enhancing the model's generalization on mechanical parts.}

\item We design a novel conditioning framework that aligns 2D image-derived point clouds with CAD sequence encodings via contrastive learning. This enables our VQ-Diffusion model to predict topologically valid CAD sequences, which are subsequently compiled into standard STEP files for downstream tasks.

\end{itemize}

\section{Related Work}
\label{sec:rw}

\subsection{Parametric CAD Representation}

Representation in CAD generation must bridge geometric precision and generative flexibility. Early approaches relied on standard BRep or parametric surfaces (e.g., Bézier, NURBS)~\cite{piegl2012nurbs}, which are geometrically exact but possess non-Euclidean structures notoriously difficult for neural networks to process. DeepCAD~\cite{wu2021deepcad} shifted this paradigm by abstracting CAD models into sequential operations (e.g., sketch, extrude), enabling autoregressive Transformer modeling.

Recent advancements span three streams. First, direct BRep generation enforces topological consistency: BrepGPT~\cite{li2025brepgpt} uses Voronoi Half-Patches autoregressively, while BrepDiff~\cite{lee2025brepdiff} adapts diffusion via masked UV grids. \changed{By directly generating BReps, such approaches can explore a wider topological space.} Second, logic-enhanced methods like Boolean CAD~\cite{2025Boolean} use Constructive Solid Geometry (CSG)~\cite{sharma2018csgnet} to resolve complex Boolean combinations. Third, LLM-based works like CAD-Llama~\cite{li2025cad}, \changed{CAD-MLLM~\cite{xu2024cad}}, and FlexCAD~\cite{Zhang2025FlexCAD} convert parametric sequences into editable code scripts.

Despite these advances, methods often trade sequence learnability for topological validity. Autoregressive sequences become prohibitively long, while direct BRep generation struggles with global coherence. In contrast, we restructure the sequence-based paradigm through a coarse-to-fine lens. Our hierarchical three-level codebook compresses operations into a discrete latent space, generating fully parametric, topologically robust models without LLM computational overhead.

\subsection{Reverse Engineering}
Reverse engineering geometric inputs (e.g., point clouds) into CAD models remains challenging. Traditional analytical methods~\cite{roth1993extracting,xia2020geometric} for primitive segmentation and surface fitting demand manual intervention and lack watertight guarantees. Geometric Deep Learning~\cite{atz2021geometric,bronstein2017geometric,cao2020comprehensive} automates this. TransCAD~\cite{dupont2024transcad} pioneered predicting modeling sequences directly from point clouds via Transformers. AutoBRep~\cite{xu2025autobrep} reconstructs BRep topology by learning surface adjacencies. Addressing manufacturing artifacts, DeFillet~\cite{jiang2025defillet} removes fillets to recover sharp edges. Recently, CAD-Recode~\cite{rukhovich2025cad} utilized LLMs to decode point clouds into executable Python scripts, automating ``scan-to-code''.

However, current methods bifurcate into two extremes: traditional fitting lacks editability, while LLM-based approaches are resource-intensive and noise-sensitive. Our work bridges this gap by automating ``geometry-to-sequence'' extraction focused on \changed{heuristic-guided resampling}~\cite{xie2025iovs4nerf}. Specifically, our UA-DGCNN selectively resamples points to preserve sharp edges and smooth surfaces. This tailored approach effectively filters noise from intermediate geometry, enabling precise modeling sequence recovery.

\subsection{Image-Driven CAD Reconstruction}

Single-view image reconstruction is a prominent frontier. While neural rendering (e.g., NeRF) and diffusion-based mesh generators (Wonder3D~\cite{long2024wonder3d}, TripoSR~\cite{tochilkin2024triposr}) produce impressive visual geometries, their outputs lack the clear internal CAD topology of curves and surfaces.

To achieve CAD generation, recent works utilize intermediate proxies or direct BRep generation. CADDreamer~\cite{li2025caddreamer} generates normal maps via 2D diffusion priors, followed by geometric optimization to recover BReps. Img2CAD~\cite{you2025img2cad} uses Vision-Language Models (VLMs) to factorize images into SVGs, inferring 3D commands via code. \changed{Concurrently, GenCAD~\cite{alam2024gencad} aligns visual features with CAD sequences via contrastive pretraining for diffusion generation, while CADCrafter~\cite{chen2025cadcrafter} employs Diffusion Transformers (DiT) in a structured latent space.} Similarly, HoLa~\cite{liu2025hola} unifies geometry and topology in a holistic latent space.

However, these methods face domain adaptation and information loss bottlenecks. First, synthetic datasets like ShapeNet dominate, which lack manufacturable shapes and realistic renders. Second, \changed{approaches like CADDreamer and Img2CAD rely on 2D priors (normal maps or SVGs)} to infer 3D structures, inherently losing depth and topology. This makes reconstruction ill-posed and prone to hallucinations.

To resolve the data gap, \changed{we utilize CAD-220K, a curated subset of the ABC dataset~\cite{Koch_2019_CVPR}, alongside PrintCAD, a collection of 3D-printed solids.} To bridge the modality gap, we upgrade the intermediate representation to 3D point clouds to preserve spatial structure. Crucially, rather than using them merely as inputs, we employ contrastive learning to map image features directly to the CAD sequence space. This ensures the generation of structurally aligned, STEP-standard CAD models from a single image.

\section{Methodology}
\label{sec:method}

As illustrated in Fig.~\ref{fig:pipeline}, the Img2CADSeq framework addresses the ill-posed single-view reconstruction problem via a structured, three-stage pipeline designed to progressively resolve geometric ambiguity.

First, Hierarchical Sequence Encoding. We introduce a three-level sequence encoder. Driven by the prior that shape recognition proceeds from ``global structure to local details'', we design a hierarchical three-level codebook. Unlike flat representations, we decouple global semantics from local geometry, encoding CAD sequences into a compact, discrete latent space via Vector Quantized Variational Autoencoders (VQ-VAE)~\cite{van2017neural}.

Second, Geometry-Aware Feature Extraction. To bridge the domain gap, we lift the single-view image into a 3D point cloud. To mitigate noise and redundancy, a UA-DGCNN refines this intermediate geometry, extracting robust feature embeddings that encode essential topological cues.

Third, Cross-Modal Alignment and Generation. We employ a contrastive learning framework to align the point cloud embeddings with the CAD latent space. These aligned features serve as structural conditions for a VQ-Diffusion model. Finally, a geometric kernel compiles the predicted tokens into a BRep model.

\begin{figure*}
  \centering
  \includegraphics[width=0.9\linewidth]{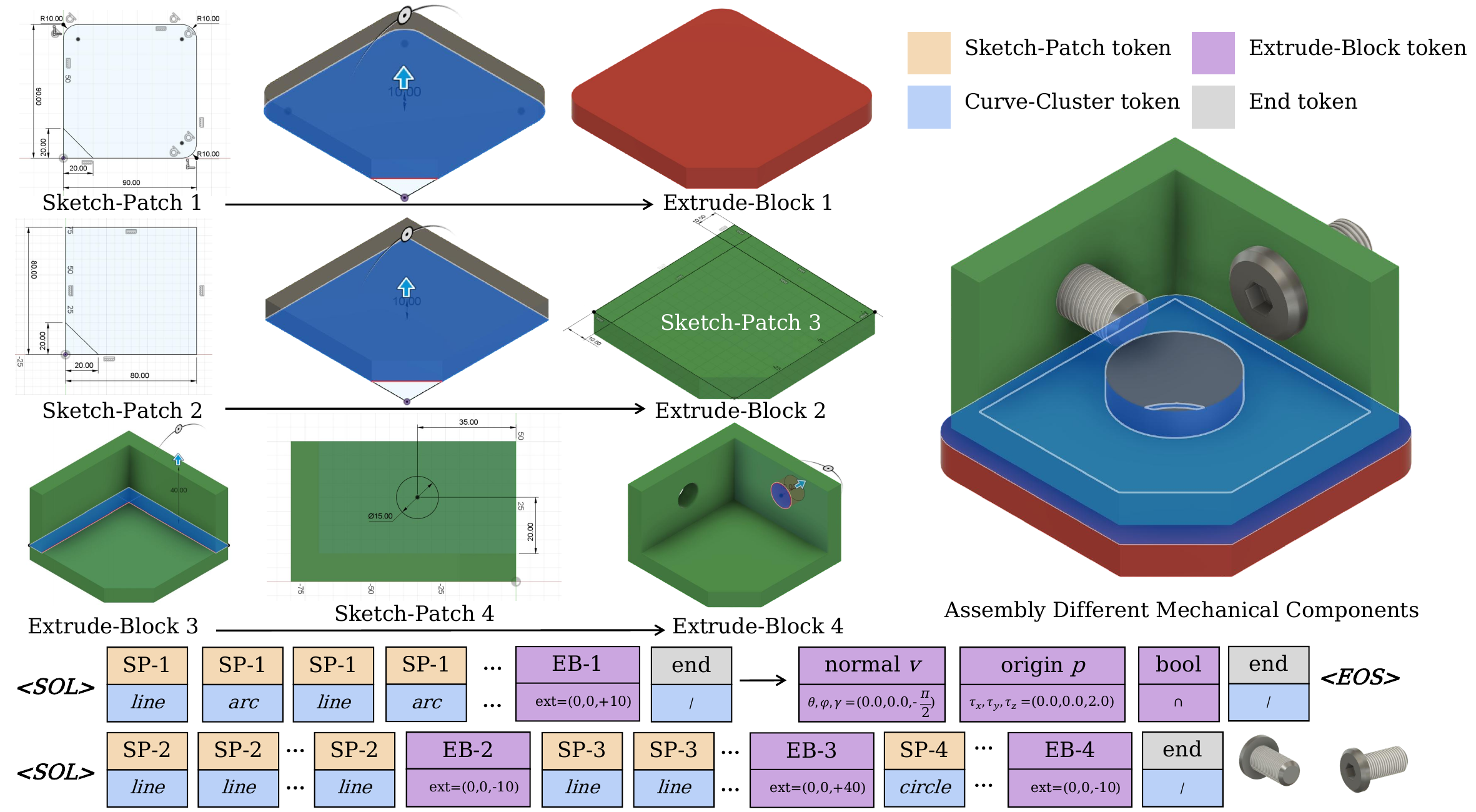}
  \caption{
  \textbf{Workflow of Hierarchical Entity Construction.}  At the base level, the \textbf{Curve-Cluster} parameterizes geometric primitives, which form closed loops in the \textbf{Sketch-Patch}. These loops are then lifted into 3D space via a normal vector and origin to perform extrusion and Boolean operations, resulting in an \textbf{Extrude-Block}. Multiple blocks are finally assembled to yield the target solid. This process mirrors the construction history of standard CAD workflows, preserving human design intent.
  }
  \label{fig:sae}
\end{figure*}

\subsection{CAD Sequence Encoder}

A fundamental bottleneck in CAD generation lies in the spatial-semantic coupling inherent in raw operation sequences. \changed{While sharing the goal of sequence-based generation with recent methods like SkexGen~\cite{xu2022skexgen}, we avoid parallel encoders that limit vertical hierarchical dependencies. Building upon the hierarchical VQ-VAE paradigm of HNC-CAD~\cite{xu2023hierarchical}, we address its over-reliance on absolute Euclidean coordinates}—a formulation that entangles local geometry with global placement and violates translation invariance, thus inhibiting transferable geometric
primitives.

To address this, we introduce a three-level codebook \changed{that utilizes a novel sorting mechanism and a local coordinate formulation.} The modeling process is factorized into: Extrude-Block (EB) for global semantics, Sketch-Patch (SP) for topological layout, and Curve-Cluster (CC) for local geometry. (see Fig. \ref{fig:sae})

The EB layer abstracts modeling operations into parametric primitives. We encode the construction parameters into a global vector $\mathbf{e}^{\text{eb}} \in \mathbb{R}^{512}$ via a Multi-Layer Perceptron (MLP):
\begin{equation}
\mathbf{e}^{\text{eb}} = \text{MLP}_{\text{global}}\Bigl(
\mathbf{n}_{\text{sketch}} \oplus
\mathbf{p}_{\text{origin}} \oplus
h_{\text{ext}} \oplus
b_{\text{type}}
\Bigr),
\end{equation}
where $\oplus$ denotes concatenation, $\mathbf{n}_{\text{sketch}}, \mathbf{p}_{\text{origin}} \in \mathbb{R}^3$ are the
unit normal vector and origin of the sketch plane, $h_{\text{ext}}$ is extrusion depth, and $b_{\text{type}} \in \{0,1\}^3$ is a one-hot Boolean indicator (New, Join, Cut). The 10-dimensional raw feature is projected to the latent space $\mathbb{R}^{512}$.

The SP layer simulates structural prioritization and establishes a spatial anchor for the relative
Curve-Cluster nodes. Unlike sorting in ascending order, we sort $m$-th loop using a score:
$S_m = \underbrace{(w_m \times h_m)}_{\text{Area}} + \alpha \cdot \underbrace{\sqrt{w_m^2 + h_m^2}}_{\text{Diagonal}} - \beta \cdot \underbrace{(x_m + y_m)}_{\text{Position Penalty}}.$
With $\alpha \gg \beta$, dominant profiles (large area or diagonal scale) rank first, using position solely as a tie-breaker based on top-left coordinates $(x_m, y_m)$. 

To enable shift-invariant encoding for the subsequent CC layer, the SP level establishes a spatial anchor:
\begin{equation}
\mathbf{e}_m^{\text{sp}} = \text{MLP}_{\text{layout}}\Bigl( S_m \oplus \mathbf{p}_{\text{start}}^{(m)} \oplus \theta_{\text{start}}^{(m)} \Bigr),
\end{equation}
where $\mathbf{p}_{\text{start}}^{(m)} = (x_0, y_0)$ and $\theta_{\text{start}}^{(m)}$ denote the absolute starting coordinates and tangent angle of the first curve in Loop $L_m$.

The CC layer discards absolute coordinates in favor of a local Frenet-frame encoding. The $i$-th primitive is formulated as an 8-dimensional feature $\mathbf{f}_i$ by concatenating geometric parameters with a type indicator and mapped to a latent code:
\begin{equation}
    \mathbf{e}_i^{\text{cc}} = \text{MLP}_{\text{curve}}(\mathbf{f}_i), \quad \text{where} \quad \mathbf{f}_i = [l_i, \Delta\theta_i, \kappa_i, \delta x_i, \delta y_i] \oplus \mathbf{t}_i.
\end{equation}
Geometric parameters include chord length $l_i$, relative tangential deviation $\Delta\theta_i$, and curvature $\kappa_i$. We also explicitly model residual offsets $(\delta x_i, \delta y_i)$ to better support geometric closure. $\mathbf{t}_i$ is a one-hot vector categorizing the primitive into $\{\text{Line}, \text{Arc/Circle}, \text{EOS}\}$. By delegating loop segmentation to the SP level, we eliminate redundant \texttt{<SEP>} tokens, allowing the CC sequence to form a continuous trajectory terminated solely by \texttt{<EOS>} to improve sequence learnability.

To learn the discrete latent codes, we employ three independent Vector Quantized Variational Autoencoders (VQ-VAE) corresponding to the EB, SP, and CC levels. Each VQ-VAE adopts a symmetric Transformer-based architecture. Level-specific MLPs are utilized to project raw heterogeneous features into a unified latent space, and a Masked Skip Connection strategy is applied between the encoder $E_\phi$ and decoder $D_\psi$ to strictly enforce codebook reliance. The networks are trained by optimizing a total loss $\mathcal{L}$, which is formulated as a weighted sum of reconstruction, quantization, and geometric closure terms:
\begin{equation}
\mathcal{L} = \mathcal{L}_{\text{recon}} + \mathcal{L}_{\text{vq}} + \lambda_{\text{cls}} \mathcal{L}_{\text{closure}}.
\end{equation}
Specifically, the hybrid reconstruction loss $\mathcal{L}_{\text{recon}}$ (balanced by $\alpha=1.0, \beta=0.5$) accounts for heterogeneous data types by combining a Mean Squared Error (MSE) term for continuous geometric parameters (such as $l_i, \Delta\theta_i$ in CC or $h_{\text{ext}}$ in EB) and a Cross-Entropy (CE) term for discrete categorical indicators (such as $\mathbf{t}_i$ or $b_{\text{type}}$). The quantization loss $\mathcal{L}_{\text{vq}} = \|\text{sg}[\mathbf{z}_e] - \mathbf{z}_q\|_2^2 + 0.25 \|\mathbf{z}_e - \text{sg}[\mathbf{z}_q]\|_2^2$ employs a standard commitment mechanism to stabilize codebook updates, where $\text{sg}[\cdot]$ denotes the stop-gradient operator. Finally, to enforce the watertightness of the generated CAD models, we introduce a loop closure regularization $\mathcal{L}_{\text{closure}}$ specifically for the CC level, which penalizes the cumulative Euclidean error of the relative path reconstruction, defined as $\|\sum \mathcal{T}(l_i, \Delta\theta_i)\|_2^2$, where $\mathcal{T}$ transforms the relative intrinsic parameters back to global displacement vectors, to minimize open-loop artifacts. We select $\lambda_{\text{cls}}=1.0$ in the experiment.

\subsection{Point Cloud Acquisition}

To lift the 2D input $\mathcal{I} \in \mathbb{R}^{H \times W \times 3}$ into a 3D point cloud, we adapt the Dens3R method. To bridge the gap between generic objects and rigid industrial parts, we employ Parameter-Efficient Fine-Tuning (PEFT) on the Query and Value linear layers. We optimize projection weights $\Theta$ to align spatial attention with CAD priors:

\begin{equation}
\Theta^* = \arg\min_{\Theta} \mathcal{L}_{\text{rec}}\left( f_{}(\mathcal{I}; \mathbf{W}_{\text{frozen}}, \Theta), \mathcal{P}_{\text{GT}} \right),
\end{equation}
where $\mathbf{W}_{\text{frozen}}$ represents the frozen weights of the backbone, $\Theta$ denotes the learnable parameters of the network $f$, and $\mathcal{P}_{\text{GT}}$ refers to the corresponding ground truth 3D point cloud. By refining $\Theta$, we effectively recalibrate the reasoning of the model to align with the rigid geometric priors of CAD models, generating an initial coarse point cloud $\mathcal{P}_{\text{raw}}$ with reasonable global structures but noisy boundaries.

To refine noisy boundaries without losing sharp edges, we propose an Uncertainty-Aware DGCNN. The backbone processes 3D vertices via four sequential EdgeConv layers to predict a geometric importance score $s_i \in [0, 1]$. A high $s_i$ indicates regions of high curvature or geometric importance (e.g., sharp edges, corners). To transform these scores into a selection mask, we design a \changed{Heuristic-guided Resampling} strategy that balances feature preservation with global coverage. The probability $p(p_i)$ of selecting a point $p_i$ is formulated as a mixture distribution:
\begin{equation}
p(p_i) = \underbrace{\lambda \cdot \frac{e^{\beta s_i}}{\sum_{k} e^{\beta s_k}}}_{\text{Saliency Term}} + \underbrace{(1-\lambda) \cdot \frac{1}{N}}_{\text{Coverage Term}}.
\end{equation}
Here, the Saliency Term utilizes a Boltzmann distribution (controlled by temperature $\beta$) to aggressively prioritize high-importance points, ensuring that sharp features are almost deterministically retained. The Coverage Term ensures a uniform background sampling to prevent voids in planar regions. $\lambda$ serves as the trade-off coefficient.

Finally, we sample $M$ points based on $p(p_i)$ to form the refined point cloud $\mathcal{P}_{\text{ref}}$. This set is then projected into the latent space via the geometric encoder $\mathcal{E}$ to obtain the structural embedding $\mathbf{z}_{\mathcal{P}} = \mathcal{E}(\mathcal{P}_{\text{ref}})$, which serves as the geometry-aligned condition for the subsequent CAD sequence generation.

\subsection{Cloud–to–CAD Conversion}

 Once the geometric embedding $\mathbf{z}_{\mathcal{P}}$ is obtained from the point cloud, the critical next step is to map it into the semantic space of CAD operations. Instead of training a CAD autoencoder in isolation, we propose a joint contrastive learning framework to enforce a shared manifold between the two domains.

We formulate the CAD sequence encoder using a causal Transformer architecture. Let a CAD program $\mathcal{S}$ be represented by a sequence of discrete codebook embeddings $\mathcal{S} = (\mathbf{e}_1, \dots, \mathbf{e}_L)$, where $\mathbf{e}_i$ corresponds to the hierarchical tokens. We append learnable positional encodings to these tokens and process them through causal self-attention layers to produce the global CAD embedding $\mathbf{z}_{\mathcal{S}}$. Unlike unsupervised reconstruction, we optimize this representation to be distinct yet semantically aligned with $\mathbf{z}_{\mathcal{P}}$.

 We employ the InfoNCE loss to maximize the mutual information between matched pairs while pushing apart mismatched ones. Given a batch of $B$ synchronized pairs $\{(\mathcal{S}_i, \mathcal{P}_i)\}_{i=1}^B$, we treat the corresponding $(\mathbf{z}_{\mathcal{S}, i}, \mathbf{z}_{\mathcal{P}, i})$ as the positive pair, and the remaining $2(B-1)$ samples within the batch as negatives. The symmetric contrastive loss is defined as:
\begin{equation}
\mathcal{L}_{\text{NCE}} = - \frac{1}{B} \sum_{i=1}^B \log \frac{\exp(\text{sim}(\mathbf{z}_{\mathcal{S}, i}, \mathbf{z}_{\mathcal{P}, i}) / \tau)}{\sum_{k=1}^{2B} \mathbb{I}_{[k \neq i]} \exp(\text{sim}(\mathbf{z}_{\mathcal{S}, i}, \mathbf{z}_{\text{feat}, k}) / \tau)},
\end{equation}
where $\text{sim}(\mathbf{u}, \mathbf{v}) = \mathbf{u}^\top \mathbf{v} / (\|\mathbf{u}\| \|\mathbf{v}\|)$ denotes cosine similarity, $\tau$ is the learnable temperature, and $\mathbf{z}_{\text{feat}} \in \{\mathbf{z}_{\mathcal{S}}\} \cup \{\mathbf{z}_{\mathcal{P}}\}$ represents the set of all embeddings in the batch.

With the aligned condition $\mathbf{c} = \mathbf{z}_{\mathcal{P}}$ established, we frame the CAD generation as a conditional discrete diffusion process. The target CAD operation sequence is first quantized into discrete indices $\mathbf{x}_0 \in \{1, \dots, K\}^L$ via our codebook.
We adopt a VQ-Diffusion framework utilizing an absorbing state transition. The forward process $q(\mathbf{x}_t | \mathbf{x}_{t-1})$ progressively corrupts the sequence by replacing tokens with a generic [MASK] token. The reverse process $p_\theta(\mathbf{x}_{t-1} | \mathbf{x}_t, \mathbf{c})$ aims to recover the clean topology $\mathbf{x}_0$ from the noisy state $\mathbf{x}_t$, explicitly guided by the structural condition $\mathbf{c}$.

We parameterize the denoising network to predict the probability distribution of the original token $\mathbf{x}_0$ directly at each timestep $t$:
\begin{equation}
p_\theta(\mathbf{x}_{t-1} | \mathbf{x}_t, \mathbf{c}) \propto \sum_{\tilde{\mathbf{x}}_0} q(\mathbf{x}_{t-1} | \mathbf{x}_t, \tilde{\mathbf{x}}_0) p_\theta(\tilde{\mathbf{x}}_0 | \mathbf{x}_t, \mathbf{c}),
\label{eq:reverse_process}
\end{equation}
where $p_\theta(\tilde{\mathbf{x}}_0 | \mathbf{x}_t, \mathbf{c})$ is modeled by a Transformer decoder that cross-attends to the point cloud embedding $\mathbf{c}$. The training objective combines the variational lower bound (VLB) with an auxiliary cross-entropy loss:
\begin{equation}
\mathcal{L}_{\text{gen}} = \mathcal{L}_{\text{vlb}} + \lambda_{\text{aux}} \mathbb{E}_{t, \mathbf{x}_t} \left[ -\log p_{\theta}(\mathbf{x}_{0} | \mathbf{x}_{t}, \mathbf{c}) \right].
\label{eq:loss}
\end{equation}

\section{Experiments}
\label{sec:results}

\subsection{Setup}
To ensure robust sequence learning and sim-to-real generalization, we curate distinct data subsets for our pipeline stages.

For the sequence learning, we train the hierarchical codebook on DeepCAD~\cite{wu2021deepcad}, \changed{a dataset comprising ground-truth CAD operation sequences. Filtering out compilation failures and non-manifold geometry yields 168,656 valid CAD programs with sequence lengths ranging from 3 to 59 operations. Our sequence encoder uses a standard 80/10/10 train/val/test split in line with existing baselines to ensure a fair comparison.}

\changed{To support the point cloud lifting stage, we utilize CAD-220K (Synthetic), a curated subset of the ABC dataset~\cite{Koch_2019_CVPR} filtered by surface count. Observing that models with 11–50 faces constitute the vast majority (339,489 in total), we proportionally downsample the data to establish balanced complexity tiers: 40K models (1–10 faces), 120K (11–50 faces), 30K (51–100 faces), and 30K (>100 faces). For these 220K models, we generate the corresponding STLs, point clouds, and four-view rendered images.}

\changed{To explore sim-to-real translation, we introduce PrintCAD, a collection of over 2,000 3D-printed solids. For each model, we systematically capture four views under real-world lighting. These objects exhibit manufacturing artifacts and texture noise, offering an evaluation of model robustness beyond synthetic renders.  Notably, both CAD-220K and PrintCAD are exclusively utilized to fine-tune our image-to-point-cloud module, ensuring that the point cloud embeddings align with the CAD latent space.}

For the visual rendering \changed{of our synthetic models}, we use DaVinciVisualizer to synthesize photorealistic images with randomized azimuth and elevation, ensuring diverse industrial viewpoints. Point clouds ($N=4,096$) are sampled from BReps compiled via OpenCascade. We enforce strict settings (0.001 linear deflection, 0.1 angular deflection) to first convert it into a mesh, and then use Trimesh to produce the uniform sampling of points.

\subsection{Implementation Details}

We implement Img2CADSeq using PyTorch and conduct all training on a server equipped with 8 NVIDIA A100 (40GB) GPUs. 

The three independent VQ-VAEs (for EB, SP, and CC levels) are trained on the DeepCAD dataset for 200 epochs. We use the AdamW optimizer with a global batch size of 1,024 (128 per GPU) and a cosine learning rate schedule warmed up to $1 \times 10^{-4}$. 
The contrastive model for geometry-aware feature alignment is trained for 300 epochs with a global batch size of 512 (64 per GPU) and a learning rate of $1 \times 10^{-3}$. During this stage, we employ a bootstrapping data augmentation strategy that subsamples 2,048 points from dense 4,096-point clouds during each forward pass.
The generative VQ-Diffusion Transformer is trained for 50,000 iterations with a global batch size of 2,048 (256 per GPU). We optimize the model using AdamW with a base learning rate of $1 \times 10^{-4}$ and a linear warmup of 5,000 steps. A linear noise scheduler is applied over 100 diffusion steps.

\subsection{Metrics}
We evaluate our method using a comprehensive set of standard metrics tailored to the specific objectives of each task. For image-conditioned generation, we assess geometric accuracy via Chamfer Distance (CD), evaluate structural integrity through the Ratio of Hanging Faces (HF), and measure primitive segmentation quality using Segmentation Accuracy (Seg Acc). For point cloud-conditioned generation, we quantify point-level geometry matching using Accuracy (Acc) and Completeness (Comp), while assessing primitive-level topology recovery via Precision and Recall. Finally, for unconditional generation, we evaluate distribution similarity against the reference set using Maximum Mean Discrepancy (MMD) and Jensen-Shannon Divergence (JSD). We also quantify generation diversity through Coverage (COV), Novelty (Nov), and Uniqueness (Uniq), and assess the programmatic validity of the generated sequences using the Invalid Rate (IR). Detailed definitions and implementations for all metrics are provided in the Appendix.

\subsection{Baselines}
Our work presents a comprehensive evaluation of BRep generation across three distinct settings, with the primary focus on single-view image conditioning. Point cloud conditioning and the unconditioning experiments serve as important extensions that demonstrate the versatility and robustness of our approach.

\paragraph{Image Input}

\begin{table}[htbp]
    \centering
    \caption{
        Quantitative results of image-conditional generation.
        CD is the Chamfer Distance (multiplied by \(10^2\)), HF is the Ratio of Hanging Faces, and Seg Acc is the Part Segmentation Accuracy.  Img2CAD~\cite{you2025img2cad} specializes in furniture like chairs and cabinets, so we do not showcase the results in the figure. TripoSR and Wonder3D~\cite{long2024wonder3d} are both methods that generate meshes from single-view and TripoSR generates rather better results, so we chose it to display. 
    }
    \label{tab:image_table}
    \setlength{\tabcolsep}{4.5pt} 
    \begin{tabular}{l|c|c|c}
        \toprule
        
        Method & CD ↓ & HF ↓ & Seg Acc ↑ \\
        \midrule
        TripoSR & 9.68 & 32.7 & 40.4\% \\
        HoLa & 1.82 & 3.3 & 90.6\% \\
        CADDreamer & 1.36 & 2.4 & 96.1\% \\ 
        Img2CAD & 1.37 & 3.0 & 86.3\%\\
        Wonder3D & 14.26 & 47.2 & 32.4\%\\
       Ours & \textbf{1.21} & \textbf{2.2} & \textbf{97.2\%} \\ 
        \bottomrule
    \end{tabular}
\end{table}

\begin{figure*}
  \centering
  \includegraphics[width=1\linewidth]{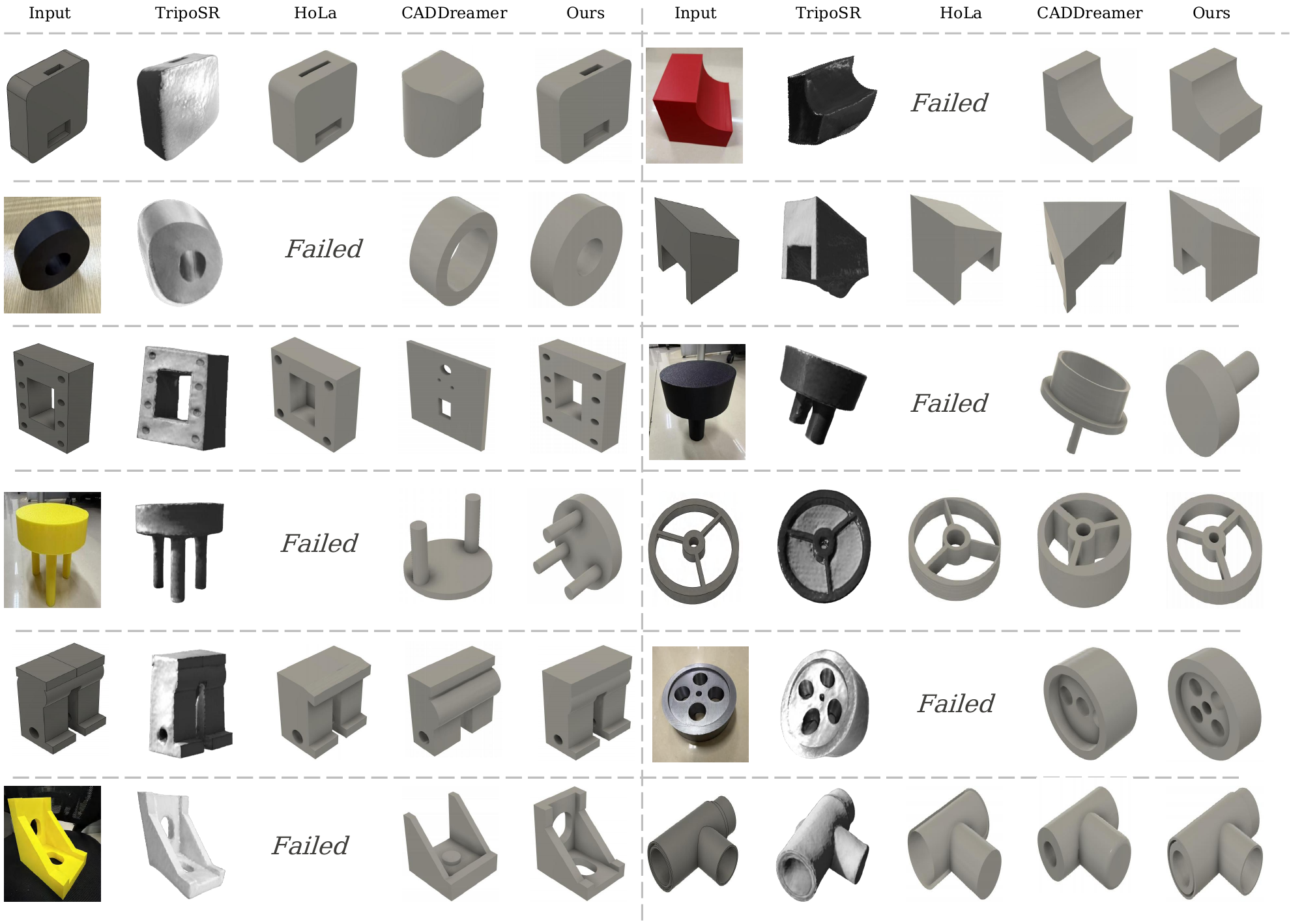}
  \caption{
  We evaluate our method on synthetic and challenging real-world images. \changed{Since DeepCAD mostly contains simple shapes where all methods perform similarly, this figure highlights complex geometries to better illustrate our advantages.} Unlike baselines that struggle with noise and smooth out sharp features, our approach reconstructs BRep models with rather cleaner topology and precise structural details.
  }
  \label{fig:image_exp}
\end{figure*}

Fig. \ref{fig:image_exp} 
visually showcases the reconstruction results from synthetic and real-world images, while Tab. \ref{tab:image_table} provides quantitative
comparisons. These results demonstrate that Img2CADSeq
offers significant advantages in generation quality and the topological fidelity of the reconstructed BReps.

Img2CADSeq’s key strength lies in its ability to simultaneously capture low-level geometric features and high-level shape understanding. Our method generates clean, compact BReps with sharp edges, minimizing fitting errors
and reconstruction failures. In contrast, competing methods generate rather distorted reconstructions. 
In addition, our method employs a refined three-level codebook and \changed{heuristic-guided resampling} for sharp edges and smooth surfaces. Single-view RGB image reconstructions often produce meshes with errors such as uneven surfaces and oversmoothed boundaries. As shown in Figure~\ref{fig:image_exp}, TripoSR~\cite{tochilkin2024triposr} often generates shapes that resemble simple clay models.
 HoLa~\cite{liu2025hola} generates a rather better BRep structure, but input ambiguity leads to unreasonable results, making them unsuitable for real-world applications. 
Since CADDreamer~\cite{li2025caddreamer} uses primitive fitting methods, it achieves higher precision on surfaces such as tori and cones, but its output cannot align well with the input, and it fails in intersecting relationships, sometimes producing non-watertight models. 
In contrast, our method demonstrates superior generation that aligns closely with the ground truth CAD models, resulting in the lowest CD, HF, and highest Seg Acc among compared approaches, confirming that our generated models are both more accurate and structurally diverse.

\paragraph{Point Cloud Input}

\begin{table}[H]
    \centering
    \caption{
        Quantitative results of clean point cloud-conditioned generation. Acc Err is the Accuracy Error, and the Comp Err is the Completeness Error. Since SEDNet is an improved version of \changed{HPNet~\cite{yan2021hpnet}}, we chose to show SEDNet in our figure, and include HPNet's results in the table. Acc Err and Comp Err scores are multiplied by \(10^3\).
    }
    \label{tab:pointcloud_table}
    \setlength{\tabcolsep}{4.5pt} 
    \begin{tabular}{l|c|c|c|c}
        \toprule
        Method & Acc Err ↓ & Comp Err ↓ & Precision ↑ & Recall ↑\\
        \midrule
        HoLa & 6.59 & 5.05 & 0.944 & 0.920\\ 
        HP+P2C & 12.27 & 49.50 & 0.832&0.708\\
        SED+P2C & 7.60& 47.10 & 0.860 & 0.676 \\
       Ours & \textbf{6.49} & \textbf{4.91} & \textbf{0.947} &\textbf{0.923} \\ 
        \bottomrule
    \end{tabular}
\end{table}

\begin{table}[H]
    \centering
    \caption{
        Quantitative comparison of unconditional CAD generation. MMD is the Maximum Mean Discrepancy, JSD is the Jensen-Shannon Divergence, COV is the Coverage, Nov is the Novelty, Uniq is the Uniqueness, and IR is the Invalid Ratio. JSD scores are multiplied by \(10^2\).
    }
    \label{tab:uncondition-table}
    \setlength{\tabcolsep}{3.5pt} 
    \begin{tabular}{l|c|c|c|c|c|c}
        \toprule
        Method & MMD ↓ & Nov ↑ & COV ↑ & Uniq ↑ & JSD ↓ & IR ↓ \\
        \midrule
        SkexGen &1.480 & 93.9\% & 73.61\% & \textbf{97.8\%} & 1.129 & 11.23 \\
        HNC-CAD & 1.296 & 94.7\% & 73.86\% & 90.7\% & 1.484 & 12.09 \\
        DTGBrepGen & 1.083 & \textbf{98.8\%} & 74.29\% & 95.8\% & 1.284 & \textbf{9.57} \\
        BrepDiff & 1.320 & 93.2\% & 75.21\% & 91.8\% & 1.127 & 11.76 \\
        HoLa & 1.069 & 92.7\% & 78.87\% & 93.3\% & 1.109 & 17.32 \\
        Ours & \textbf{0.958} & 97.2\% & \textbf{80.17\%} & 96.4\% & \textbf{0.839} & 10.89 \\ 
        \bottomrule
    \end{tabular}
\end{table}

\begin{figure*}
  \centering
  \includegraphics[width=1\linewidth]{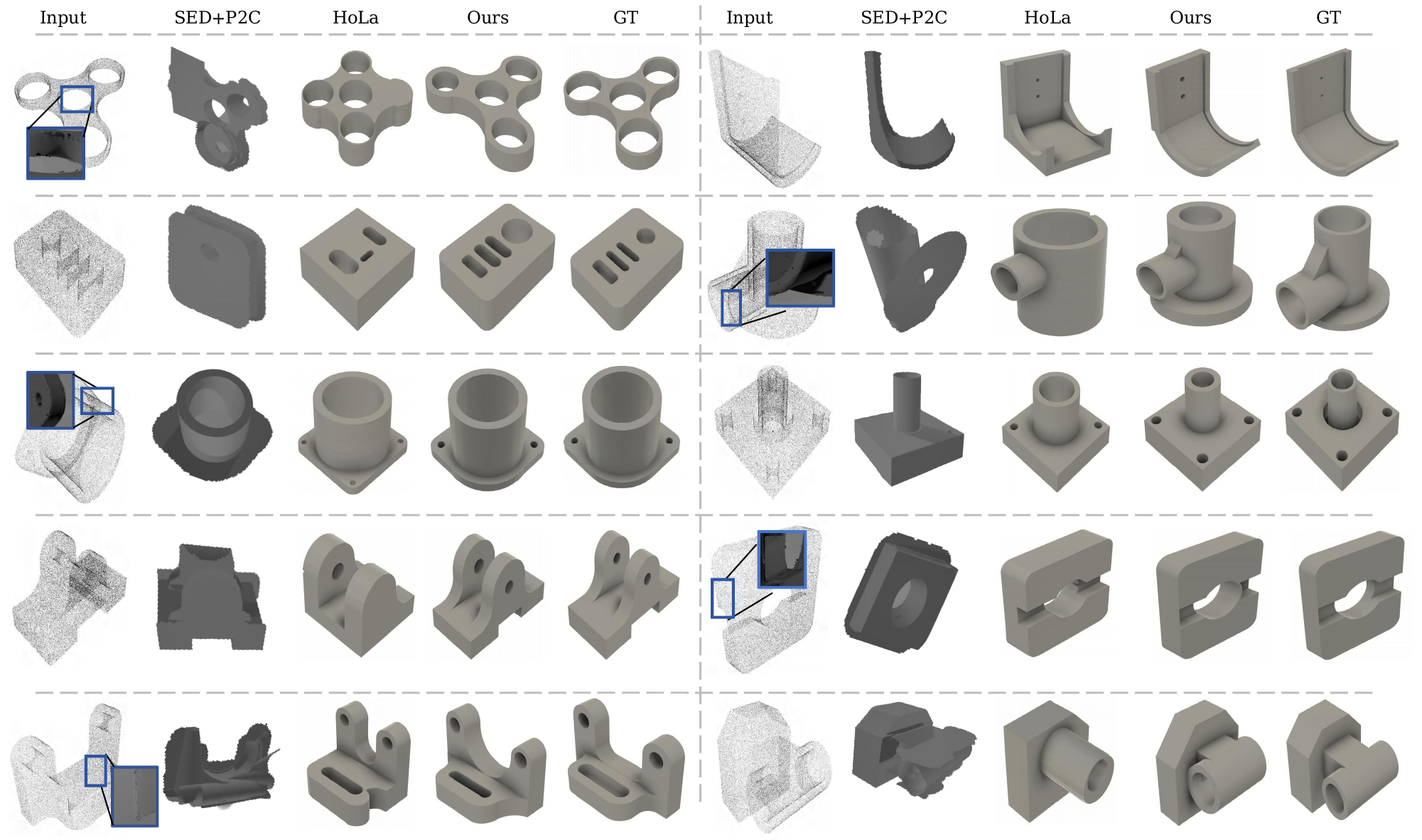}
  \caption{
  We evaluate our method against state-of-the-art approaches on inputs with ill-scanned point clouds with misalignment parts, or the clean ones.
  While baseline methods often produce distorted shapes or fail to recover topology under heavy noise, our approach can still synthesize multiple plausible CAD candidates.
  }
  \label{fig:pointcloud_exp}
\end{figure*}

Fig. \ref{fig:pointcloud_exp} showcases the reconstruction results
from clean and ill-scanned point clouds, while Tab. \ref{tab:pointcloud_table} provides quantitative comparisons.
Visually, our method consistently produces clean, watertight BRep models that preserve sharp edges and fine geometric features. In contrast, fitting-based methods like \changed{SEDNet~\cite{li2023surface}}+Point2CAD\cite{liu2024point2cad} tend to produce more complete surfaces, and the lack of topological information prevents these surfaces from being correctly trimmed. While a major limitation of HoLa is its parametrically uneditable files, the images and generated results still cannot be properly aligned. Our method leverages a more robust codebook and noisy-friendly representation that captures
both geometry and topology, supported by our method’s leading performance in Tab. \ref{tab:pointcloud_table}, where it achieves the lowest Acc Err, Comp Err and the highest precision and recall scores, demonstrating its ability to cover diverse shape modes and generate a wide variety of structurally valid CAD candidates even from partial or noisy inputs.

\paragraph{Unconditional Generation}

\begin{figure*}
  \centering
  \includegraphics[width=1\linewidth]{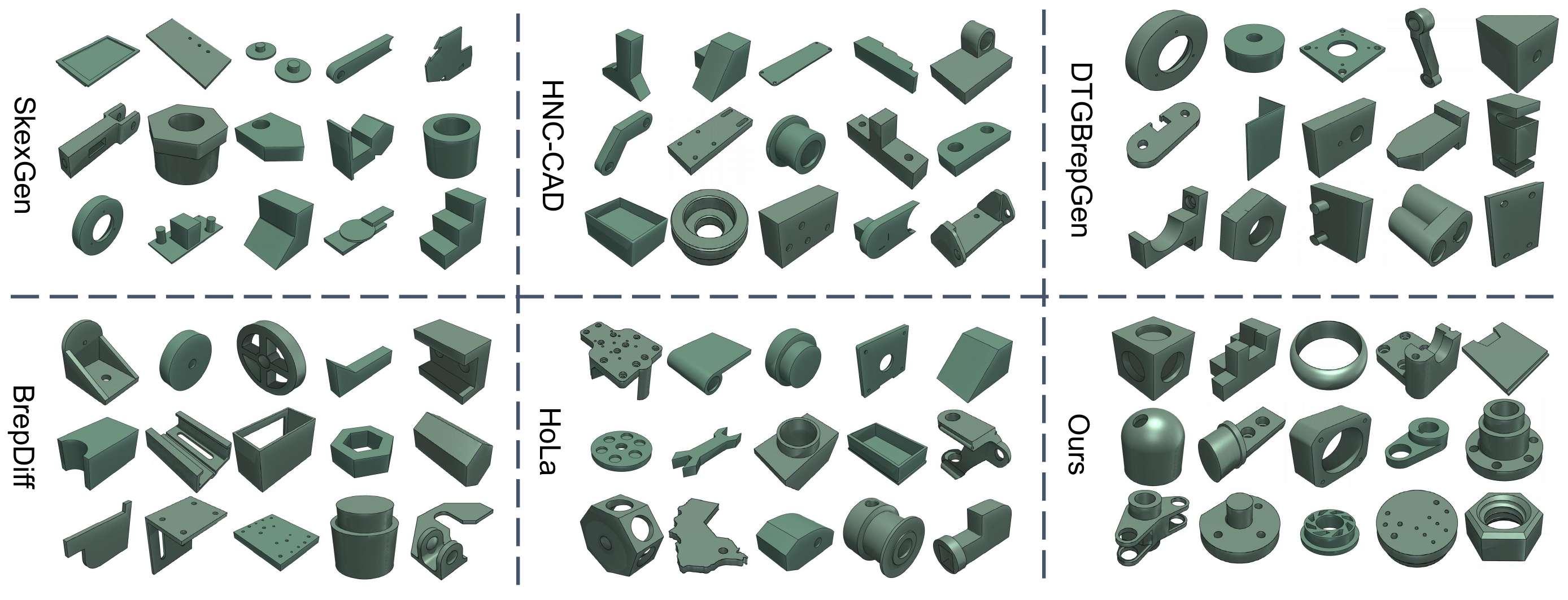}
  \caption{
   We compare our Img2CADSeq with other widely adopted baselines in unconditional generation.
   Our method produces structurally plausible models with clean surfaces and coherent mechanical features, while baseline methods often exhibit simplistic shapes, poor assembly, or unnatural proportions, demonstrating our clear advantage in generating production-ready CAD candidates.
  }
  \label{fig:uncondition}
\end{figure*}

Fig. \ref{fig:uncondition} presents a direct comparison of unconditionally generated CAD models, and Tab. \ref{tab:uncondition-table} provides quantitative comparisons. Our method produces models that are more structurally plausible, with cleaner surfaces and more coherent mechanical features. In contrast, baseline methods such as SkexGen\cite{xu2022skexgen}, HNC-CAD~\cite{xu2023hierarchical}, and DTGBrepGen~\cite{li2025dtgbrepgen} often generate rather simplistic or poorly assembled shapes, while BrepDiff~\cite{lee2025brepdiff} and HoLa exhibit surface artifacts or unnatural proportions. Our results demonstrate a clear advantage in generating production-ready CAD candidates.

\begin{figure}
  \centering
  \includegraphics[width=1\linewidth]{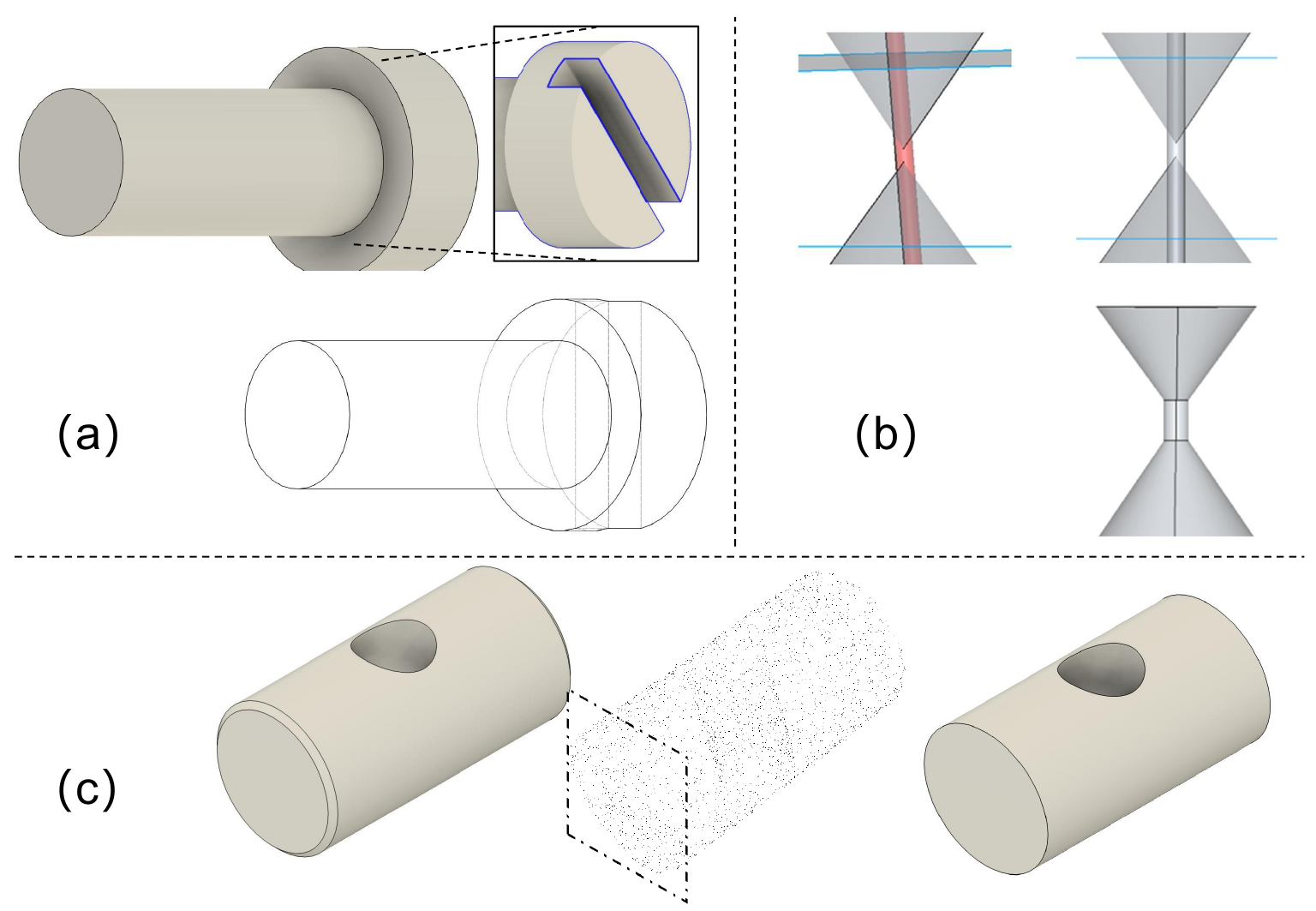}
  \caption{
  \textbf{The limitations and failure cases of our work.} (a) Extreme single-view ambiguity causes plausible back-end structures but non-manufacturable in occluded areas. (b)Sequence error accumulation disrupts global geometric constraints like strict symmetry and coaxiality. (c) Limited resolution of the intermediate point cloud causes fine features to be smoothed out or omitted.
  }
  \label{fig:fc}
\end{figure}

\section{Limitations and Failure Cases}
While our method advances CAD reconstruction, several limitations remain. For complex assemblies, long sequences cause accumulated errors. Additionally, while our PrintCAD dataset supports physical object reconstruction, it focuses mostly on 3D-printed materials, making it difficult to generalize to highly diverse in-the-wild scenarios. In practice, these constraints manifest in three main failure cases. First, as illustrated in Fig. \ref{fig:fc}(a), due to single-view ambiguity, the model hallucinates occluded regions based on learned priors. While topologically watertight, these generated back-end structures often lack physical and kinematic constraints; under severe occlusion, they may appear visually plausible but are physically impossible to manufacture. Second, as demonstrated in Fig. \ref{fig:fc}(b), although our codebook captures local features effectively, it lacks explicit global constraints. Consequently, sequence error accumulation easily disrupts strict symmetry and coaxiality, and enforcing such global constraints in diffusion models remains a major challenge. Finally, as shown in Fig. \ref{fig:fc}(c), bridging the modality gap with a limited-resolution point cloud proxy (e.g., 4,096 points) often smooths out micro-structures like threads and tiny fillets during heuristic-guided resampling, causing the final CAD sequences to miss these precise manufacturing details.

\section{Conclusions and Future Work}
\label{sec:conclusion}

We introduced Img2CADSeq, a multi-stage pipeline that directly generates topologically valid STEP files from single views. Our hierarchical codebook and contrastive alignment—powered by the new CAD-220K and PrintCAD datasets—successfully decouple global structure from local geometry, outperforming baselines in handling the sim-to-real gap. To further resolve single-view ambiguity, future work will integrate VLMs for multimodal conditioning, enable interactive editing at intermediate stages, and explicitly enforce strict symmetry and geometric constraints. By delivering editable BReps rather than inert meshes, this work significantly advances automated reverse engineering and intelligent downstream manufacturing.

\begin{acks}
The authors thank the anonymous reviewers for their valuable feedback. This work was supported by the Shenzhen Innovation and Entrepreneurship Plan under grant numbers 20232910020 and KJZD20230923114114028.
\end{acks}

\clearpage
\bibliographystyle{ACM-Reference-Format}
\bibliography{src/ref}

\clearpage

\begin{figure*}
  \centering
  \includegraphics[width=1\linewidth]{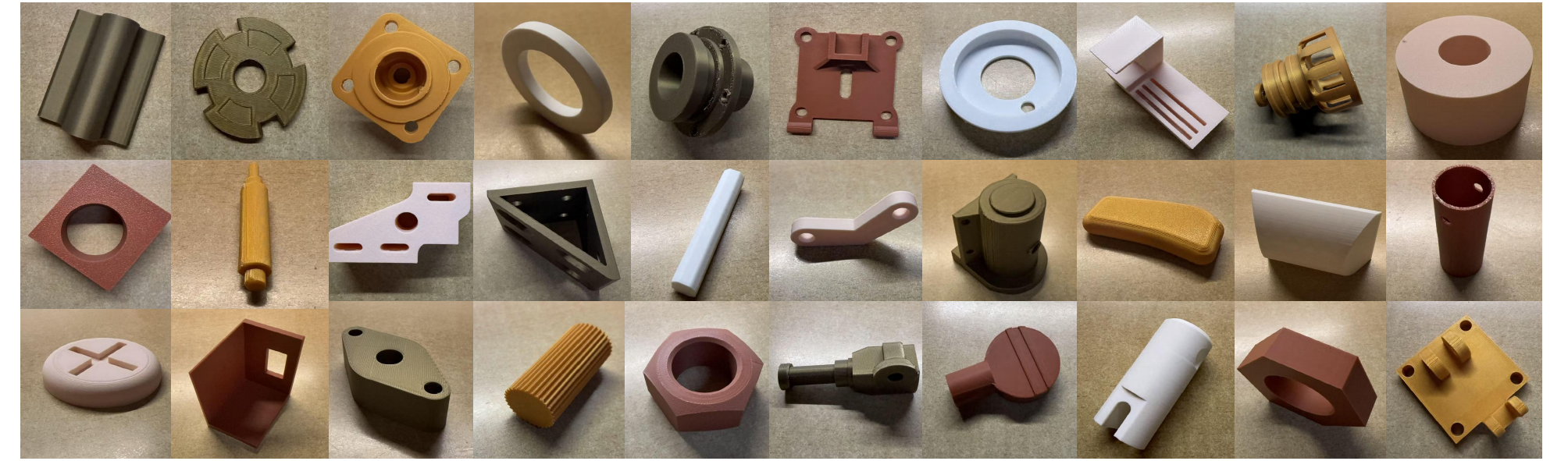}
  \caption{
Here are some samples from our newly introduced dataset, PrintCAD, which comprises over 2,000 3D printed objects captured under uncontrolled real-world lighting conditions using an iPhone. The dataset aligns real-world images with corresponding ground-truth CAD models, spanning a variety of materials (Nylon, Resin, PLA) and geometric complexities.
  }
  \label{fig:dataset}
\end{figure*}

\clearpage

\clearpage
\appendix
\section*{SUPPLEMENTARY MATERIALS}
\renewcommand{\thesection}{\Alph{section}} 

\section{Evaluation Metrics}

We report metrics across three settings to assess three different conditions. Following standard evaluation protocols, instances where the geometric kernel fails to compile a valid STEP file are excluded from the metric computation. Meanwhile, following BrepGen, we generate 3,000 valid models, randomly sample 1,000 instances from the test set 10 separate times, and report the averaged results.

\subsection{Image-Conditional Metrics}
\begin{itemize}
    \item \textbf{Chamfer Distance (CD):} The average minimum Euclidean distance between 10,000 uniformly sampled points on the reconstructed model and the ground truth.
    \item \textbf{Ratio of Hanging Faces (HF):} The percentage of faces with open edges (not shared with neighbors), serving as a proxy for topological watertightness.
    \item \textbf{Segmentation Accuracy (Seg Acc):} The percentage of surface points whose primitive type labels correctly match the ground truth.
\end{itemize}

\subsection{Point Cloud-Conditional Metrics}
\begin{itemize}
    \item \textbf{Accuracy Error (Acc Err) \& Completeness Error (Comp Err):} The mean one-way distance from generated points to ground truth (Acc Err), and from ground truth to generated points (Comp Err).
    \item \textbf{Precision \& Recall:} The percentage of generated primitives that match ground truth primitives (Precision), and the percentage of ground truth primitives successfully recovered (Recall), within a distance threshold of 0.1.
\end{itemize}

\subsection{Unconditional Generation Metrics}
\begin{itemize}
    \item \textbf{Maximum Mean Discrepancy (MMD):} The average CD between generated samples and their nearest neighbors in the test set, quantifying geometric fidelity.
    \item \textbf{Coverage (COV):} The percentage of test set samples matched by at least one generated sample, quantifying mode coverage.
    \item \textbf{Jensen-Shannon Divergence (JSD):} The divergence between the voxelized distributions ($28^3$ grid) of generated and reference sets.
    \item \textbf{Invalid Rate (IR):} The percentage of generated sequences that fail to compile into valid BRep geometry.
    \item \textbf{Novelty (Nov):} The proportion of generated samples that are geometrically distinct from the training set.
    \item \textbf{Uniqueness (Uniq):} The proportion of non-duplicate samples within the generated batch itself.
\end{itemize}

\section{Ablation Study}
Our method benefits primarily from three distinct technical contributions: a hierarchical sequence encoding strategy, a geometry-aware point cloud bridge, and an industrial-specific domain adaptation process. We conduct ablation studies on these contributions to investigate the performance gains achieved independently by each module. We compare our full model against four variants:

\begin{itemize}
    \item \textbf{Model 1 (Encoder with HNC-CAD):} We remove the ``overall profiles to local details'' sorting mechanism in the SP level and the relative encoding in the CC level, reverting to a standard flat sequence representation similar to HNC-CAD.
   \item \textbf{Model 2 (Encoder with SkexGen):} While sharing the goal of sequence-based generation, we replace our hierarchical structure with the parallel encoder architecture of SkexGen, which completely decouples topology and geometry. This variant removes our novel sorting mechanism and local coordinate formulations.
    \item \textbf{Model 3 (w/o Point Cloud Bridge):} We bypass the intermediate point cloud acquisition and cross-modal alignment. The 2D image features (extracted by a ViT) are directly fed into the VQ-Diffusion model to predict the CAD sequence.
    \item \textbf{Model 4 (w/o Domain Adaptation):} We use the original base model without our fine-tuning on CAD-220K and PrintCAD datasets, and we replace our heuristic-guided resampling with standard random sampling.
\end{itemize}

\begin{table}[H]
    \centering
    \caption{\textbf{Ablation Study on Core Components.} 
    Quantitative ablation results. We report the ratio of Hanging Faces (HF), Chamfer Distance (CD), and Segmentation Accuracy (Seg Acc). Best values are highlighted in bold.
    }
    \label{tab:ablation}
    \setlength{\tabcolsep}{5pt}
    \begin{tabular}{l|c|c|c}
        \toprule
        Config & CD ($\times 10^{2}$) $\downarrow$ & HF (\%) $\downarrow$ & Seg Acc (\%) $\uparrow$ \\
        \midrule
        Model 1 & 1.65 & 14.8 & 91.5 \\
        Model 2 & 3.42 & 10.2 & 88.5 \\
        Model 3 & 4.12 & 6.4 & 84.3 \\
        Model 4 & 2.45 & 5.1 & 93.8 \\
        \midrule
        \textbf{Full Model} & \textbf{1.21} & \textbf{2.2} & \textbf{97.2} \\
        \bottomrule
    \end{tabular}
\end{table}

First, we evaluate the contribution of our hierarchical sequence encoding strategy by comparing our three-level codebook against alternative representations. Tab.~\ref{tab:ablation} presents the performance results of Models 1 and 2. As shown, removing our specific hierarchical design in Model 1 leads to a sharp increase in Hanging Faces (HF) from 2.2\% to 14.8\%. This confirms that without the explicit canonical ordering and relative coordinate system, the autoregressive model struggles to maintain loop closure and topological validity. Similarly, Model 2 yields significantly higher Chamfer Distance (CD) and HF compared to our full approach. This indicates that our vertical hierarchical dependencies---from global profiles to local details---are more effective at preserving structural coherence.

Next, we investigate the necessity of our intermediate 3D representation through Model 3, which eliminates the point cloud acquisition and cross-modal alignment. Since Model 3 directly predicts CAD sequences from 2D images, we focus our analysis on its geometric fidelity. As detailed in Tab.~\ref{tab:ablation}, Model 3 results in a significant performance degradation, yielding the highest CD (4.12) and significantly lower Segmentation Accuracy (84.3\%). This degradation highlights the challenge of directly bridging the semantic gap between 2D pixels and CAD operations. The point cloud bridge acts as a critical geometric anchor, explicitly resolving depth ambiguity and providing the structured condition necessary for accurate sequence generation.

Finally, we evaluate the impact of our industrial-specific tuning by analyzing Model 4, which removes our domain adaptation and heuristic-guided resampling. While Model 4 captures the global shape reasonably well (achieving a lower HF than Models 1 and 2), non-negligible geometric errors persist, with CD and Seg Acc metrics trailing our full model. Without fine-tuning on CAD-220K and PrintCAD, the network relies on generic shape priors and fails to reconstruct the sharp mechanical edges characteristic of manufactured parts. Moreover, the absence of heuristic-guided resampling limits the recovery of micro-structures, leading to over-smoothed corners. This highlights the vital role of our data and resampling strategies in robust CAD reconstruction.

\section{More Experimental Results}
\begin{figure*}
  \centering
  \includegraphics[width=\linewidth]{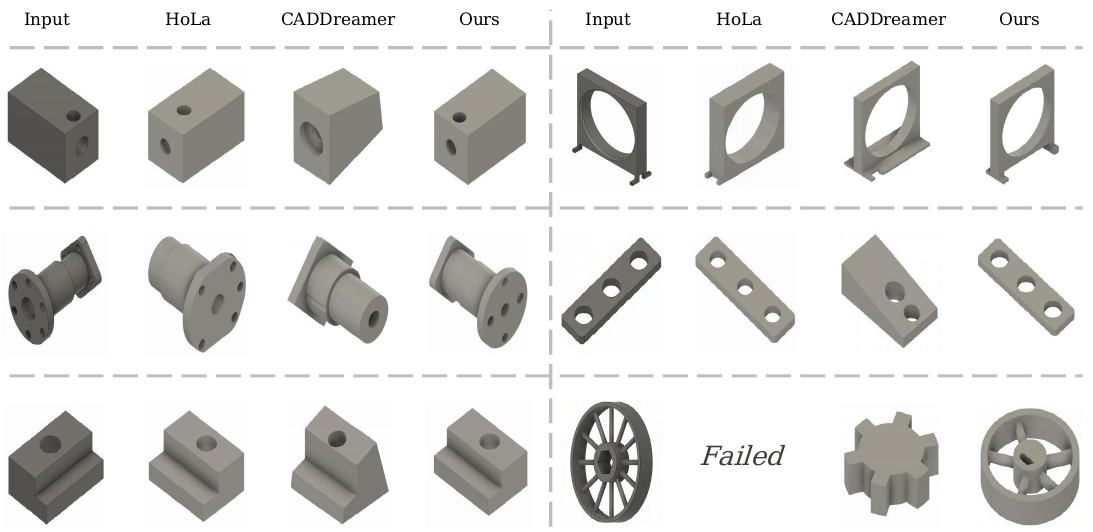}
  \caption{
\changed{Random test set samples. While simple geometries are easily handled by most methods, increasing complexity challenges all approaches. Despite limitations in extreme cases, our method better preserves global shape and visual consistency, demonstrating robustness without selection bias.}
  }
  \label{fig:re1}
\end{figure*}

We present additional, randomly selected reconstruction results in Fig.~\ref{fig:re1}. 
Current CAD datasets, such as DeepCAD, predominantly feature simpler geometries where most existing methods perform adequately. However, as face count and topological complexity increase, the generative challenge escalates for all sequence-based models. In these high-complexity scenarios, all existing methods, including ours, face inherent generative difficulties. Nonetheless, even when processing highly intricate parts, Img2CADSeq demonstrates better preservation of the global outer shape and essential geometric features compared to the baselines.

\begin{figure*}
  \centering
  \includegraphics[width=\linewidth]{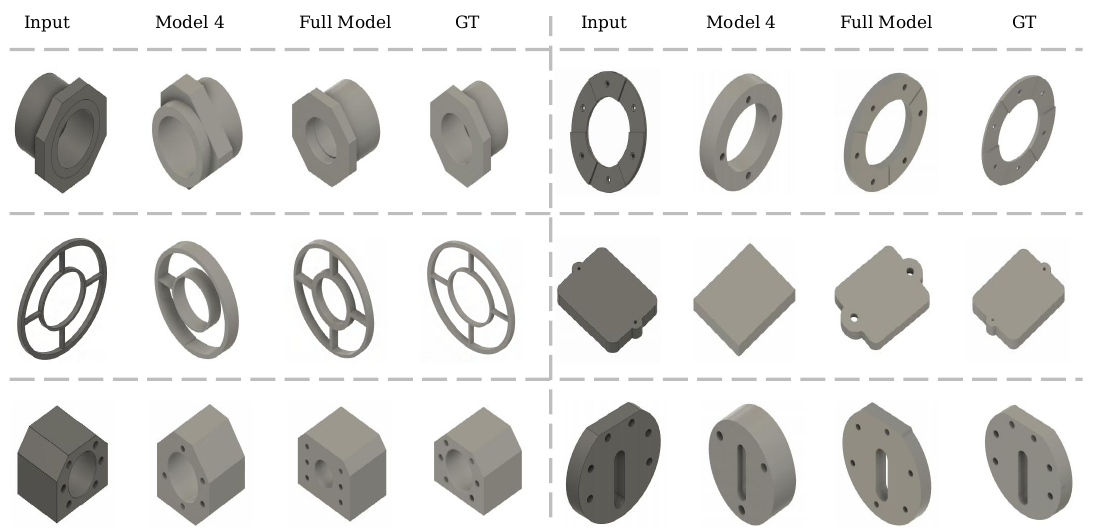}
  \caption{
\changed{Visual ablation study on fine-tuning with the two datasets. Model 4 exhibits severe geometric distortions and missing features. Our full model captures these complex structural details better, demonstrating the necessity of the extra data for generating industrial-grade CAD models.}
  }
  \label{fig:re2}
\end{figure*}

We also present a visual comparison to demonstrate the critical role of the combination of two data types during the fine-tuning stage. Fig.~\ref{fig:re2} illustrates the reconstruction results when the model is trained with and without the additional fine-tuning data. 

As observed in the Model 4 column, the model struggles to capture complex industrial features when relying solely on the base dataset. For instance, in the first row, Model 4 fails to reconstruct one of the four holes. In the second row, it completely misses the semi-circular side slots. In contrast, our full model reconstructs these intricate details better, closely aligning with the Ground Truth.

\begin{figure*}
  \centering
  \includegraphics[width=\linewidth]{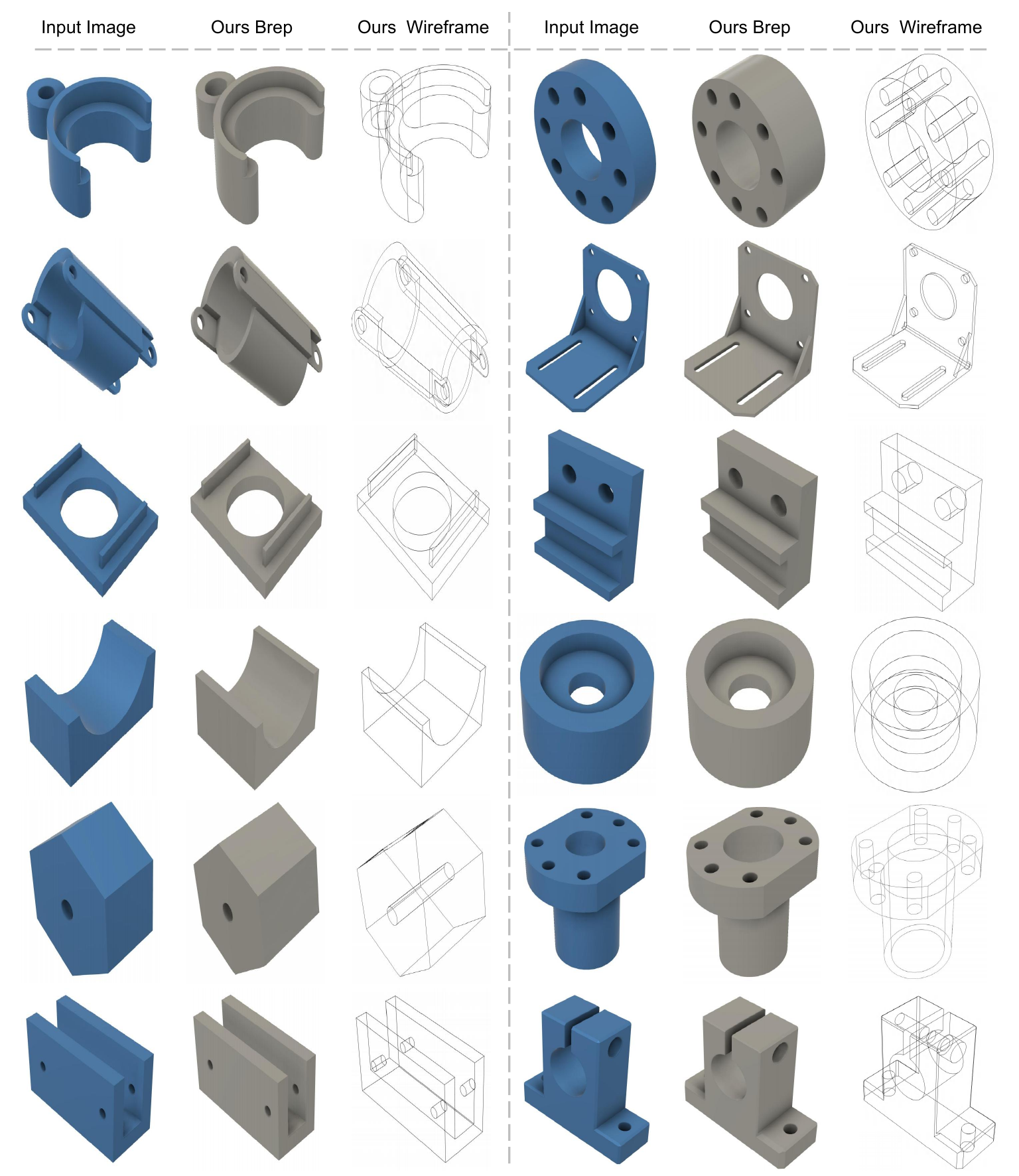}
  \caption{
Reconstruction results from the given images are shown from left to right: input image, BReps, and their CAD vertices and edges. 
  }
  \label{fig:sup1}
\end{figure*}

\begin{figure*}
  \centering
  \includegraphics[width=\linewidth]{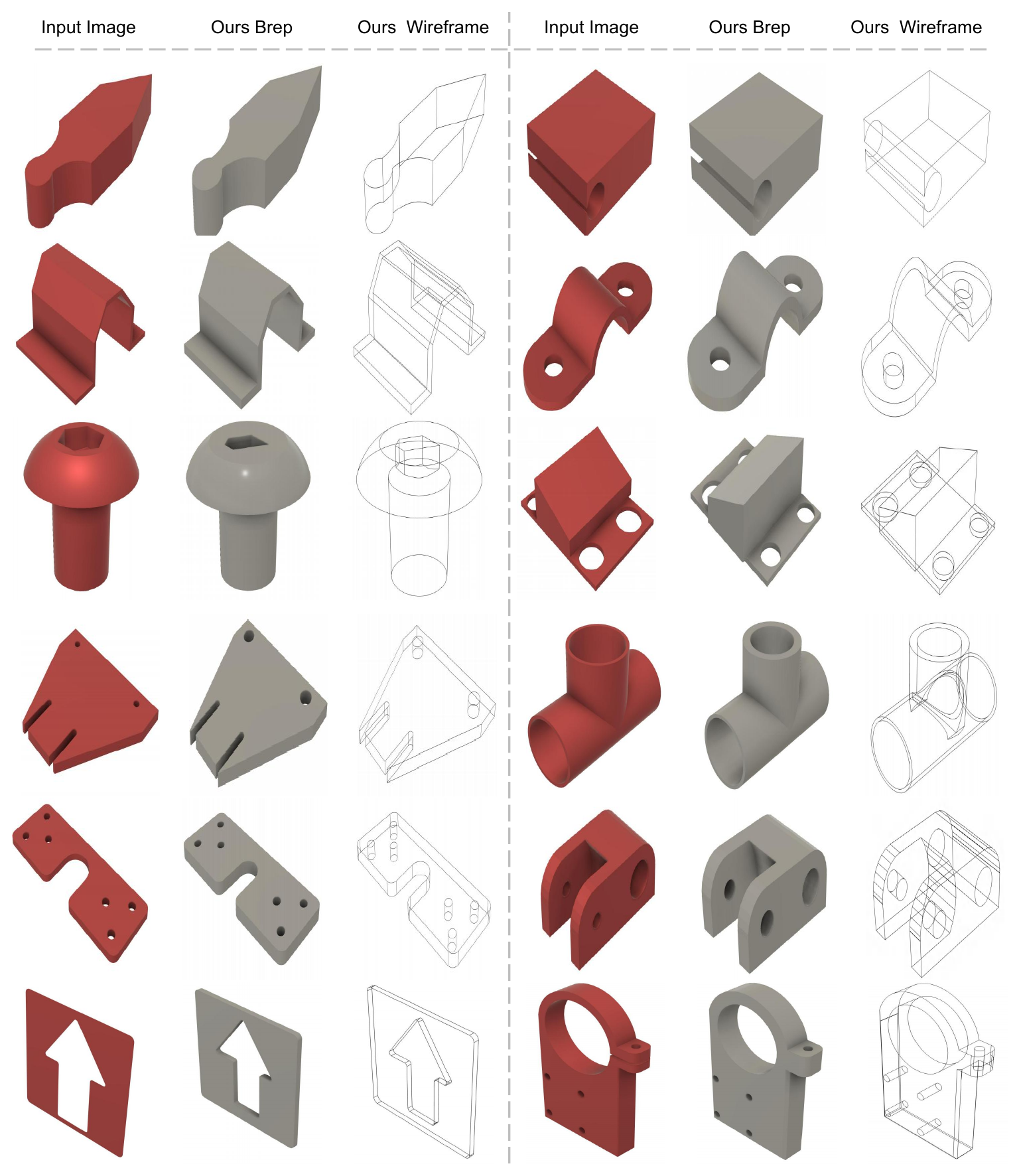}
  \caption{
Reconstruction results from the given images are shown from left to right: input image, BReps, and their CAD vertices and edges. 
  }
  \label{fig:sup2}
\end{figure*}

Fig.~\ref{fig:sup1} and Fig.~\ref{fig:sup2} showcase more reconstruction results of
our method, including input images, 
reconstructed BReps, as well as their CAD wireframe, including vertices and edges.

\end{document}

%% file: src/macros.tex
\usepackage{xcolor}
\usepackage{graphicx}
\usepackage{booktabs}
\usepackage{multirow}
\usepackage{caption}
\usepackage{subcaption}
\usepackage{algorithm}
\usepackage{algpseudocode}
\usepackage{tikz}

\definecolor{reviewblue}{rgb}{0,0,0.8}
\definecolor{reviewred}{rgb}{0.8,0,0}
\definecolor{reviewgreen}{rgb}{0,0.5,0}

\newcommand{\changed}[1]{{\color{black}#1}}